\documentclass[ oneside,journal]{llncs}  
\pdfoutput=1
\usepackage{geometry}                		
\usepackage{graphicx}				
\usepackage{amssymb}
\usepackage{amsmath}
\usepackage{adjustbox}
\usepackage{caption}
\usepackage{soul}
\usepackage{multicol}
\usepackage{calc}
\usepackage{floatrow}

\graphicspath{ {images/} }

\makeatletter
\newcommand*\@dblLabelI {}
\newcommand*\@dblLabelII {}
\newcommand*\@dblequationAux {}

\def\@dblequationAux #1,#2,%
    {\def\@dblLabelI{\label{#1}}\def\@dblLabelII{\label{#2}}}

\newcommand*{\doubleequation}[3][]{%
    \par\vskip\abovedisplayskip\noindent
    \if\relax\detokenize{#1}\relax
       \let\@dblLabelI\@empty
       \let\@dblLabelII\@empty
    \else 
       \@dblequationAux #1,%
    \fi
    \makebox[0.5\linewidth-1.5em]{%
     \hspace{\stretch2}%
     \makebox[0pt]{$\displaystyle #2$}%
     \hspace{\stretch1}%
    }%
    \makebox[0.5\linewidth-1.5em]{%
     \hspace{\stretch1}%
     \makebox[0pt]{$\displaystyle #3$}%
     \hspace{\stretch2}%
    }%
    \makebox[3em][r]{(%
  \refstepcounter{equation}\theequation\@dblLabelI, 
  \refstepcounter{equation}\theequation\@dblLabelII)}%
  \par\vskip\belowdisplayskip
}
\makeatother

\title{The RNN-ELM Classifier}
\author{Athanasios Vlontzos}
\institute{Department of Electrical and
Electronic Engineering
Imperial College
London, UK
}

\date{July 2016}							

\begin{document}
\maketitle
\begin{abstract}
 In this paper we examine  learning methods combining the Random Neural Network, a biologically inspired neural network and the Extreme Learning Machine that achieve state of the art classification performance while requiring much shorter training time.
The Random Neural Network is a integrate and fire computational model of a neural network whose mathematical structure permits the efficient analysis of large ensembles of neurons. An activation function is derived from the RNN and used in an Extreme Learning Machine. We compare the performance of this combination against the ELM with various activation functions, we reduce the input dimensionality via PCA and compare its performance \textit{vs.} autoencoder based versions of the RNN-ELM. 
 \end{abstract}

\section{Introduction}
Deep Learning, using convolutional neural networks with multiple layers of hidden units has in recent years achieved human-competitive or even better than human performance in image classification tasks \cite{Ciresan1},\cite{Hinton2012} at the expense of long training times and specialised hardware \cite{Ciresan2}. In this paper we combine the Random Neural Network (RNN)\cite{Gelenbe89},\cite{rnn},\cite{rnn-2} and the Extreme Learning Machine (ELM)\cite{Huang2006} in shallow and deep classifiers and compare their performance. 
\paragraph{The RNN:} The RNN  is a stochastic integer state, integrate and fire system \cite{dlrnn}, initially developed to model biological neurons \cite{biological} and extended to model soma-to-soma interactions \cite{soma}. It consists of M interconnected neurons, each of which can receive positive (excitatory) or negative (inhibitory) signals from external sources such as sensory sources or other cells. 
\\ The RNN can be described by equations that are possible to be solved analytically. It provides useful mathematical properties and algorithmic efficiency as seen in \cite{dlrnn} : \begin{itemize}
  \item  The state of each neuron $i$ is represented at a given time $t$ by a integer ${k}_{i} \ge 0$ which can describe the neuron's level of excitation.
  \item Each neuron $i$ receives excitatory and inhibitory spikes in the form of independent Poisson processes of rate ${ \lambda  }_{ i } ^{ + }$  and $  \lambda _ i  ^ -$. A neuron when excited (i.e. ${k}_{i} > 0$) can fire after a delay characterised by exponential distribution whose average value ${  \mu  }_{  i}^{  -1}$ depends on the specific neuron.
  \item A neuron $j$ which fires, sends an excitatory or inhibitory spike to a neuron i with probability ${ p }_{  ji}^{  +},{ p  }_{  ji}^{  -}$. We write $  w_{ j i}^+=r_j p _{  ji}^  + $ and $ w_{ j i}^-=r_j p_ {ji}^- $
  \item The state of the system is the joint probability distribution $p(k,t)= Prob[k_{1}(t),...k_{n}(t)=(k_{1},..,k_{n})]$ and it satisfies a coupled system of Chapman-Kolmogorov equations
  \item The RNN has a ``product form'' solution \cite{Gelenbe89}, meaning that in steady state, the joint probability distribution of network state is equal to the product of marginal probabilities

\end{itemize}
where the marginal :
\begin{equation}
\lim_{t\to\infty} Pr[k_i(t)=k_i]=q_i^{k_i}(1-q_i)
\end{equation}
and
\begin{equation}
q_i={{\lambda_i^+\sum_{j=1}^nq_jw_{ij}^+ }\over{r_i+\lambda_i^-\sum_{j=1}^nq_jw_{ji}^- }}
\end{equation}
The RNN was initially developed to model biological neurons \cite{biological} and has been used for landmine detection\cite{landmine1},\cite{landmine2}, video and image processing \cite{imageproc1},\cite{imageproc2},\cite{imageproc3}, combinatorial optimisation \cite{optimization}, network routing\cite{routing} and emergency management\ cite{emergency}.
\paragraph{The ELM:} The Extreme Learning Machine \cite{Huang2006} is a Single Layer Feedforward Network (SLFN) with one layer of hidden neurons. Input weights $\mathbf{W_1}$ to the hidden neurons are assigned randomly in the range [0,1] and never changed while the output weights $\mathbf{W_2}$ are estimated in one step by observing that its output is calculated as in eq. 3 where $\zeta$ is the hidden neuron activation function. Then: 
\doubleequation[eq:X,eq:Y]
{\mathbf{Y=W_2\zeta(W_1x)}}
{\mathbf{W_2=\zeta(W_1x)^+Y}}
where a least squares fit to $\mathbf{Y}$ is calculated and $()^+$ is the Moore-Penrose pseudo-inverse.\\
ELMs have been shown to achieve very good classification results and with their one-step weight estimation procedure, achieve very fast learning times. However,  ELMs tend to produce good results when very large numbers of hidden neurons are used thus reducing their computational complexity advantage since the computation time is dominated by the calculation of the pseudo-inverse of a very large matrix.

\section{The RNN-ELM and the PCA-RNN-ELM}
\paragraph{RNN-ELM:}Inspired by the fact that in mammalian brains, among other communication mechanisms, cells exhibit a quasi-simultaneous firing pattern through soma-to-soma interactions\cite{soma}, in \cite{dlrnn} an extension of the RNN was presented. A special network was considered that contained $n$ identical connected neurons, each having a firing rate $r$ and external excitatory and inhibitory spikes are denoted by ${ \lambda }^{ + }$ and ${ \lambda }^{ - }$. The state of each neuron was denoted by $q$ and each neuron receives an inhibitory input from some external neuron $u$ which is not part of the cluster, thus any cell $i$ inside the cluster has an inhibitory weight ${ w }_{ u }^{ - }\equiv { w }_{ u,i }^{ - }>0$ from the external neuron $u$ to $i$. Also the internal spiking rate weights were set to zero $w_{ i,j }^+=w_{ i,j }^-=0$. Whenever one of the neurons fires it triggers the firing of the others  with $p(i,j)=\frac { p }{ n } $ and $Q(i,j)=\frac { (1-p) }{ n } $. In this way instead of exiting or inhibiting other neurons in the cluster through spikes, the packed neurons excite each other and provoke firing through soma-to-soma interactions.
The result that \cite{dlrnn} has reached is : 
\begin{equation}
q=\frac { { \lambda  }^{ + }+\frac { rq(n-1)(1-p) }{ n-qp(n-1) }  }{ r+{ \lambda  }^{ - }+{ q }_{ u }{ w }_{ u }^{ - }+\cfrac { rqp(n-1) }{ n-qp(n-1) }  } 
\end{equation} 
which is a second degree polynomial in q that can be solved for its positive root which is the only of interest since q is a probability. 
\begin{equation}
{ q }^{ 2 }p(n-1)[{ \lambda  }^{ - }+{ q }_{ u }{ w }_{ u }^{ - }]-q(n-1)[r(1-p)-{ \lambda  }^{ + }p]\\
+n[{ \lambda  }^{ + }-r-{ \lambda  }^{ - }-{ q }_{ u }{ w }_{ u }]=0
\end{equation} 
From the standard method of solving quadratic equation we can define the activation function of the cth cluster as:

\begin{equation}\label{eq:zeta}
\zeta_C(x) = \frac {-{ b }_{ c }}{ 2{ p }_{ c }(n-1) [{ \lambda  }_{ c } ^{ - }+x]}  \\+\frac{\sqrt { { { b }_{ c } }^{ 2 }-4{ p }_{ c }(n-1)[{ \lambda  }^{-}+x]n[{ { \lambda  }_{ c } }^{ + }-{ r }_{ c }-{ { \lambda  }_{ c } }^{ - }-x] }  }{ 2{ p }_{ c }(n-1)[{ { \lambda  }_{ c } }^{ - }+x] } 
\end{equation}
where:
\begin{equation}\label{eq:zeta1}
x=\sum _{ u=1 }^{ U }{ { w }_{ u,c }^{ - } } { \overset { \_  }{ q }  }_{ u }
\end{equation}
The RNN-ELM therefore is defined as an ELM using equations (\ref{eq:zeta}) and (\ref{eq:zeta1}) as the activation function of the hidden neurons.

\paragraph{An Update Rule for ELM Output:}
In \cite{dlrnn} and \cite {dlrnn2} to achieve better accuracies in classification tasks an update rule was introduced. Instead of updating the ELM output weights based on the desired output, the desired output itself was updated and the weights were updated via the Moore-Penrose pseudo inverse. \\
 Denoting the labels of the dataset as $L={ [l_{ 1 }l_{ 2 }...l_{ D }] }^{ T }$ and the desired output as a $D\times K$ matrix , $Y=[{ y }_{ d,k }]$ where the $l_{d}$th element in $[{ { y }_{ d,1 }...y }_{ d,K }]$is initially 1 while the rest of the are set to 0. Then the hidden-layer output is then $H=\zeta (X{ W }^{ (1) })$ where $W^{(1)}$ denotes the randomly generated input weights while let $W^{(2)}$ which is determined by \begin{equation}{ W }^{ (2) }=H^+Y\end{equation}
\\ Then the output of the ELM is $O=H{ W }^{ (2) }$. 
\\The rule dictates an iterative approach to adjust Y based on the output O using the negative log-likelihood function at the cost function:
\begin{equation}
{ f }_{ d }=-ln(\frac { { e }^{ O(d,{ l }_{ d }) } }{ \sum _{ k=1 }^{ K }{ { e }^{ O(d,k) } }  } )
\end{equation}
then taking the partial derivative: 
\begin{equation}
\frac { \partial{ f }_{ d } }{ \partial O(d,k) } =
\begin{cases}
\frac { 1 }{ \sum _{ \hat { k } =1 }^{ M }{ { e }^{ O(d,\hat { k } )-O(k,d) } }  } -1\quad  & \text{if $k={ l }_{ d }$}\\

\frac { 1 }{ \sum _{ \hat { k } =1 }^{ K }{ { e }^{ O(d,\hat { k } )-O(k,d) } }  } \quad & \text{if $k \neq { l }_{ d }$}
\end{cases}
\end{equation}
then:
\begin{equation}\label{eq:update}
{ Y }^{ (i+1) }(d,k)={ O }^{ (i) }(d,k)-s\frac { \partial{ f }_{ d } }{ \partial{ O }^{ (i) }(d,k) } 
\end{equation}
where ${ O }^{ (i) }$ denotes the output after the i-th iteration based on ${ Y }^{ (i) }$ and $s>0$ is the step size chosen by the user.

\paragraph{The PCA-RNN-ELM algorithm:} 
The PCA algorithm is using Singular Value Decomposition (SVD), in that we decompose the input $\mathbf{X}$ and its covariance matrix as : 
\doubleequation[]{\mathbf{X=U\Gamma { V }^{ T }}}
{\mathbf{C=\frac { 1 }{ N } X{ X }^{ T }=\frac { 1 }{ N } U{ \Gamma  }^{ 2 }{ U }^{ T }}}
 Where $\mathbf{U}$ is an $N \times N$ matrix,  $\mathbf{\Gamma}$ is a $N\times M$ matrix and $\mathbf{V}$ is a $M\times M$ matrix.  Comparing the factorisation of $\mathbf{X}$ with that of $\mathbf{C}$ we conclude that the right vectors  $\mathbf{U}$ are equivalent to the eigenvectors of $\mathbf{XX^T}$. So the transformed data are denoted as $\mathbf{Y}$ and after selecting only the the eigenvectors corresponding to the $m$ largest eigenvalues the data are denoted as $\mathbf{Y_m}$. Both can be expressed as 
 \doubleequation[]
 {\mathbf{Y=XV= U\Gamma}}
 {\mathbf{Y_m=U_m\Gamma_m=XV_m}}Based on the above the complete PCA-ELM algorithm  with $N_h$ hidden neurons,  $\mathbf{V_m}$ the matrix of the first m principal components and training set $\mathbf{X}$ is:

\doubleequation[]
 	{\mathbf{X'_m=XV_m }}  
	  {\mathbf{W_1=rand(N_h,M)}}
\doubleequation[]
	{\mathbf{H}=\zeta(\mathbf{W_1X'_m})} 
	{\mathbf{W_2=H^+T }}

where $\mathbf{T}$ is the matrix of target labels.
Finally repeat for $I$ iterations starting with $\mathbf{O}=\mathbf{HW_2}$
\doubleequation[]
{
D^i=\frac{\partial{f}_{d}}{\partial O^{(i)}(d,k)} 
}
{
 Y^{(i)}(d,k)= O^{(i-1)}(d,k)-sD^i
}
\begin{equation}\mathbf{
W_2^i=H^+Y^i
}\end{equation}
\\
On the testing set $\mathbf{Z}$ the algorithm executed is:
\doubleequation[]
{\mathbf{
Z'_m=ZV_m
}}
{\mathbf{
Y=W_2\zeta(W_1Z'_m)
}}
When using the iterative output adaptation method, one must be careful to avoid model overfitting. The training accuracy rapidly converges to 100\% but the training accuracy starts decreasing. In our simulations we limit the training accuracy to 98.5\% for MNIST and 99\% for NORB dataset.


\section{Simulation Results}
Before we present our simulation results we must note that  most algorithms achieving better performance are using image pre-processing (e.g. \cite{LeCun1998}) or require much larger computational resources (e.g. \cite{Ciresan1},\cite{Hinton2012},\cite{rawelm}).
\paragraph{RNN-ELM \textit{vs.} ELM:}

The first experiment run was to compare the accuracy of the ELM with various activation functions with the RNN-ELM using the activation function of Eq.\ref{eq:zeta}. In \cite{rawelm} an ELM structure 784-15000-10 was used to achieve 97\% testing accuracy with a sigmoid activation function. In contrast, the RNN-ELM can achieve the same level of performance with 5000 output neurons. Table \ref{table:res11} provides a more detailed comparison using 1000 hidden neurons using various activation functions and the activation function of Eq.\ref{eq:zeta} (RNN). These results clearly demonstrate that the use of the RNN activation function leads to much better accuracies with a small increase in computational time.
\begin{table}[ht]
\caption{Activation Function Comparison Without PCA with 1000 Hidden Neurons}
\label{table:res11}
\begin{tabular}{|c|c|c|c|c|} 
\hline
\bf
\bf{Activation Function}  & \bf{Training Time}  & \bf{Testing Time}  & \bf{Training Accuracy} & \bf{Testing Accuracy} \\\hline
Sigmoid & 24,95 & 0,86 & 14.84\% & 14.05\% \\\hline
Sine  & 32,27 & 0,92 & 22.49\% & 14.99\% \\\hline
Hard Limit  & 29,27 & 0,64 & 11.24 & 11.35\% \\\hline
Triangular basis  & 13,05 & 0,63 & 9.87\% & 9.8\% \\\hline
Radial basis & 47,43 & 0,97 & 17.13\% & 17.02\% \\\hline
RNN  & 43,45 & 2,82 & 92.82\% & 92.39\% \\\hline

\hline

\end{tabular} 
\\

\end{table}

\paragraph{PCA-RNN-ELM:} 
\begin{figure} 
\centering
\includegraphics[width=85mm]{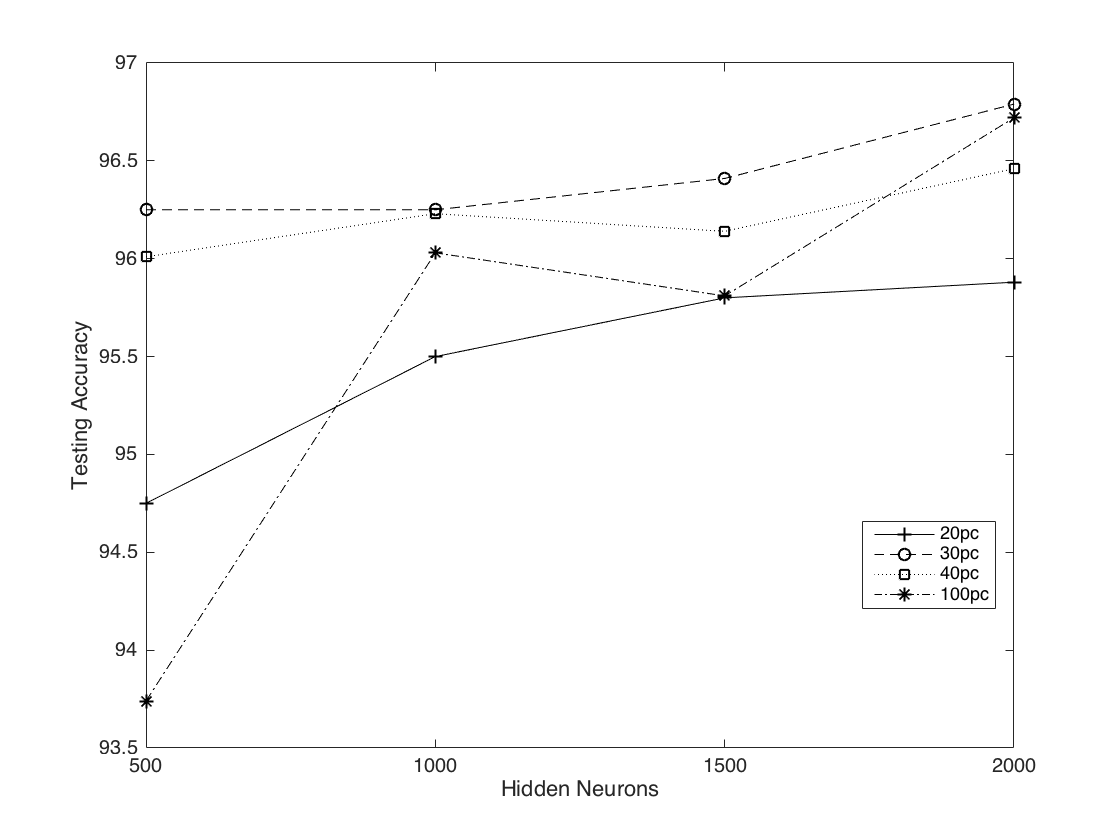}
\caption{PCA-RNN Performance for variable PCs and Hidden Neurons\label{figure}}
\centering
  \end{figure}
  \begin{table}[ht]
\caption{ MNIST Simulation results: Classifier of \cite{dlrnn} with 500-500-X structure }
\label{table:res10}
\begin{centering}

\begin{tabular}{|c|c|c|c|} 
\hline
\bf
\bf{X}   & \bf{Training Accuracy} & \bf{Testing Accuracy} &\bf{Total Time} \\ \hline
100&94.1633&94.38 & 112.9509\\\hline
200 & 96.025 & 95.82&  135.6642 \\\hline
300 & 96.515 & 96.37&  143.4844\\\hline
500&96.4517&96.21&  235.5055\\\hline
1000&96.6267& 96.48&586.1988\\\hline

\hline

\end{tabular} 
\\
\end{centering}
\end{table}

The next experiment run was to test the performance of the PCA-ELM  varying the number of principal components and the number of hidden neurons in the ELM. 
In our testing we used this process with 30 iterations and a step size of $s=5$ getting equivalent results but using less neurons as seen in figure Fig. 1.. Also its important to note that the time needed to achieve 30 iterations of the simulation seems to be constant and independent of the number of neurons used, in contrast to the method used in \cite{dlrnn} where we observe an increase in the time needed as the number of neurons increases, as seen in table \ref{table:res10}.

\paragraph{PCA-RNN-ELM \textit{vs.} Autoencoder-ELM:} We compared PCA-RNN-ELM with Autoencoder-ELM\cite{dlrnn2} using the same number of PCs and autoencoder neurons while varying the ELM size. Essentially this compares the performance of the ELM given the dimensionality reduction obtained by the two methods. Figure \ref{figure}  and Table \ref{table:res10} show that the performance of the two methods is roughly equivalent for any ELM size. The PCA-ELM enjoys a slight advantage in testing time while the accuracy is essentially the same. It must be noted that due to the randomness of the ELM weights, results vary from run to run and differences beyond the second decimal point should be ignored.
%
%
%
%
%

\paragraph{NORB with PCA-RNN-ELM:}
In testing the PCA-ELM with the NORB dataset the input data were presented to the algorithms by concatenating binocular images of the same object. Therefore, each pair of images had 2048 features (pixel values), similarly to  \cite{dlrnn}. 
We ran two sets of experiments: 1) setting ELM hidden neurons to 500 and varying the number of PCs used (Table \ref{table:res13}) and 2) keeping the number of PCs fixed and varying the number of ELM hidden neurons . Finally we ran the Deep RNN-ELM network of \cite{dlrnn} and obtained a training time of 34.88s, training accuracy 99.02\%, testing time 9.53 and testing accuracy 86.99\%.
All results are averages of 50 trials to minimise the effect of the randomised initialisation of the input weights in the ELM.
We observed also that the training times are faster than the Deep Autoencoder RNN-ELM while producing better testing accuracy by 3\% at best. The PCA-RNN-ELM always enjoys an advantage in testing time.
Given that the PCA-RNN-ELM training and testing times are quite fast, it is possible to run the algorithm on the same data multiple times and select the parameters (random input weights) that produce the best results. 
\begin{table}[ht]
\caption{NORB Simulation results: PCA-RNN-ELM with Output adaptation }
\label{table:res13}
\centering
\begin{tabular}{|c|c|c|} 
\hline
\bf
\bf{Principal Components}  & \bf{Training Accuracy} & \bf{Testing Accuracy} \\\hline
2 & 71,4856 & 59,1317 \\\hline
5  & 98,465 & 86,9753 \\\hline
10  & 99,3498 & 91,1687 \\\hline
20 &  99,2963 & 90,786 \\\hline
30 &  99,4033 & 89,856 \\\hline
40 &  99,4198 & 88,8724 \\\hline
50 &  99,2675 & 87,7202 \\\hline

\hline

\end{tabular} 
\centering

\end{table}

\section{Conclusion}
In this paper we compared the RNN-ELM to the ELM with various activation functions and observed that the RNN-ELM achieves far superior results with far fewer hidden neurons. We also demonstrated that the RNN-ELM network with PCA preprocessing is a viable alternative to other image classification algorithms by comparing three versions of the RNN-ELM network and a Deep RNN architecture on the standard MNIST and NORB datasets. The results were obtained without any prior feature extraction or image processing apart from the PCA algorithm in order to concentrate on the raw performance of the algorithms tested.
We observed that the relatively simple PCA-RNN-ELM can provide high accuracy and very fast training and testing times while the deep autoencoder-ELM algorithm can achieve similar results on the MNIST dataset.

\bibliographystyle{splncs03.bst}
\bibliography{ThanosBib}

\end{document}